\documentclass[sigconf, nonacm]{acmart}
\usepackage{subcaption}
\usepackage{graphicx}
\usepackage{multirow}
\usepackage{hyperref}
\usepackage{balance}

\AtBeginDocument{%
  }

\begin{document}

\title{OpenMoCap: Rethinking Optical Motion Capture under Real-world Occlusion}

\author{Chen Qian}
\affiliation{%
  \institution{Tsinghua University}
  \city{Beijing}
  \country{China}}
\email{chen.cronus.qian@gmail.com}

\author{Danyang Li}
\authornote{Corresponding author.}
\affiliation{%
  \institution{Tsinghua University}
  \city{Beijing}
  \country{China}}
\email{lidanyang1919@gmail.com}

\author{Xinran Yu}
\affiliation{%
  \institution{Tsinghua University}
  \city{Beijing}
  \country{China}}
\email{yuxinran0929@126.com}

\author{Zheng Yang}
\affiliation{%
  \institution{Tsinghua University}
  \city{Beijing}
  \country{China}}
\email{hmilyyz@gmail.com}

\author{Qiang Ma}
\affiliation{%
  \institution{Tsinghua University}
  \city{Beijing}
  \country{China}}
\email{tsinghuamq@gmail.com}

\renewcommand{\shortauthors}{Chen Qian, Danyang Li, Xinran Yu, Zheng Yang, and Qiang Ma.}

\begin{abstract}
Optical motion capture is a foundational technology driving advancements in cutting-edge fields such as virtual reality and film production. However, system performance suffers severely under large-scale marker occlusions common in real-world applications. An in-depth analysis identifies two primary limitations of current models: (i) the lack of training datasets accurately reflecting realistic marker occlusion patterns, and (ii) the absence of training strategies designed to capture long-range dependencies among markers. To tackle these challenges, we introduce the CMU-Occlu dataset, which incorporates ray tracing techniques to realistically simulate practical marker occlusion patterns. Furthermore, we propose OpenMoCap, a novel motion-solving model designed specifically for robust motion capture in environments with significant occlusions. Leveraging a marker-joint chain inference mechanism, OpenMoCap enables simultaneous optimization and construction of deep constraints between markers and joints. Extensive comparative experiments demonstrate that OpenMoCap consistently outperforms competing methods across diverse scenarios, while the CMU-Occlu dataset opens the door for future studies in robust motion solving. The proposed OpenMoCap is integrated into the MoSen MoCap system for practical deployment. The code is released at: \url{https://github.com/qianchen214/OpenMoCap}.
\end{abstract}





\maketitle

\section{Introduction}
\label{sec:intro}
\begin{figure}[t]
  \centering
   \includegraphics[width=0.5\linewidth]{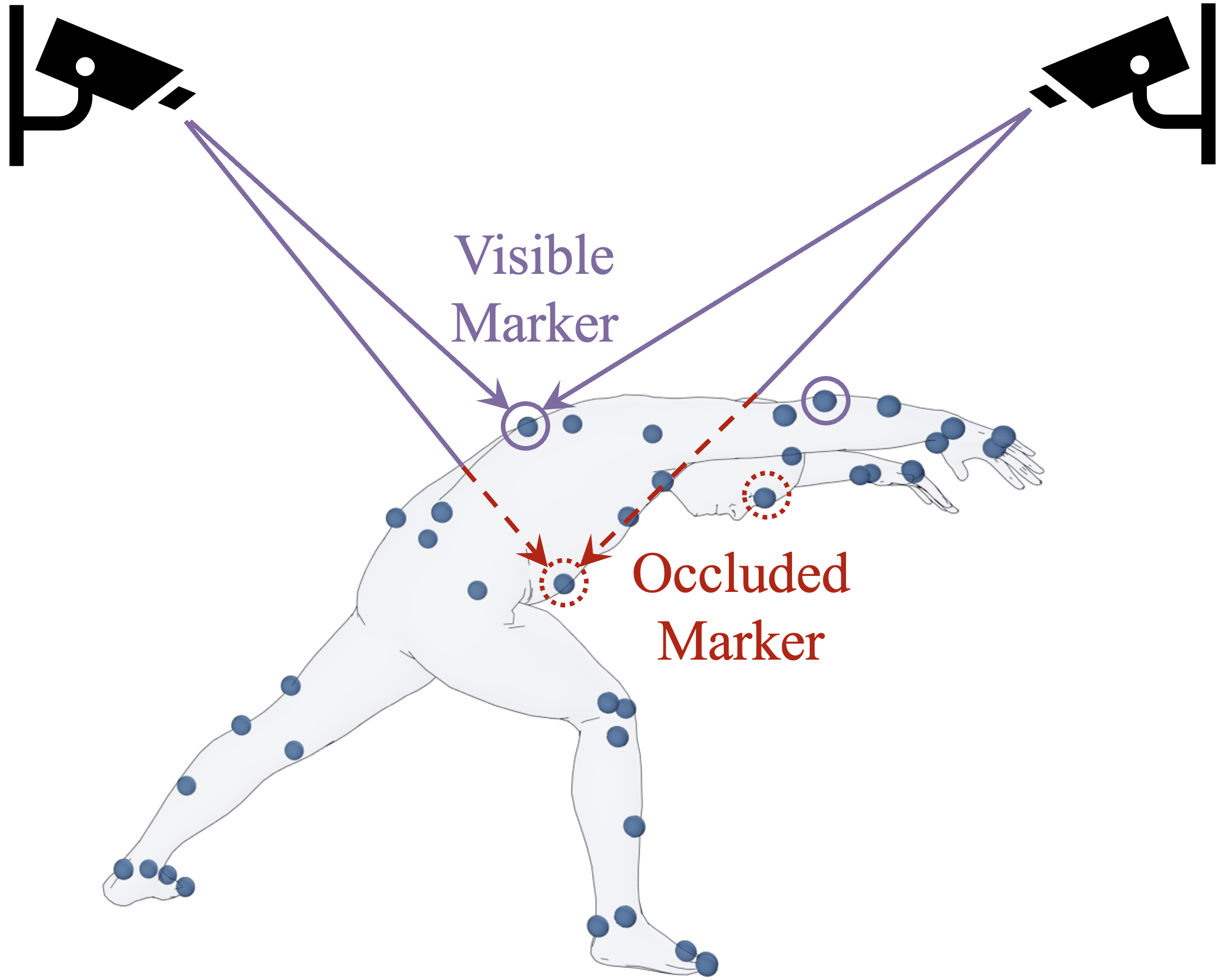}
   \caption{Marker occlusion. The marker placed on the back is captured by two infrared cameras, enabling accurate localization via triangulation. Meanwhile, the marker on the abdomen is occluded in either view.}
   \label{fig:intro_ray}
\end{figure}

Optical motion capture (MoCap) technology plays a critical role in digitally recording and reconstructing human motion patterns with high precision. This technology has become indispensable across diverse fields such as film production, game development, and embodied intelligence \cite{wang2022ocr, 
 wang2023decenternet, chatzitofis2021democap, van2018accuracy, longo2022optical}. During the MoCap process, multiple infrared cameras synchronously emit infrared light at specific wavelengths (e.g., 850 nm) and capture reflected signals from markers attached to the human body. These signals enable precise positional reconstruction for markers through triangulation methods \cite{hartley1997triangulation}. The subsequent stage, known as MoCap data solving \cite{aristidou2018self, mahmood2019amass, kim2024damo, pan2023locality}, involves deriving skeletal movements from noisy marker point clouds, thus facilitating reliable motion analysis.

Deep learning-based approaches to MoCap data solving~\cite{chen2021mocap, kim2024damo, pan2023locality, pan2024romo, mahmood2019amass} are emerging as a focal point of research in this field. MoCap-Solver~\cite{chen2021mocap} decomposes the task into three components—template skeletons, marker layout, and motion, and jointly decodes their latent representations. LocalMoCap~\cite{pan2023locality} utilizes spatially and temporally adjacent markers for mutual completion and designs graph neural networks to reconstruct human skeletons. Building upon LocalMoCap, RoMo~\cite{pan2024romo} further reduces the complexity of motion solving by decomposing joint rotations into directional components, improving both efficiency and accuracy.

\begin{figure}[t]
  \centering
   \includegraphics[width=1\linewidth]{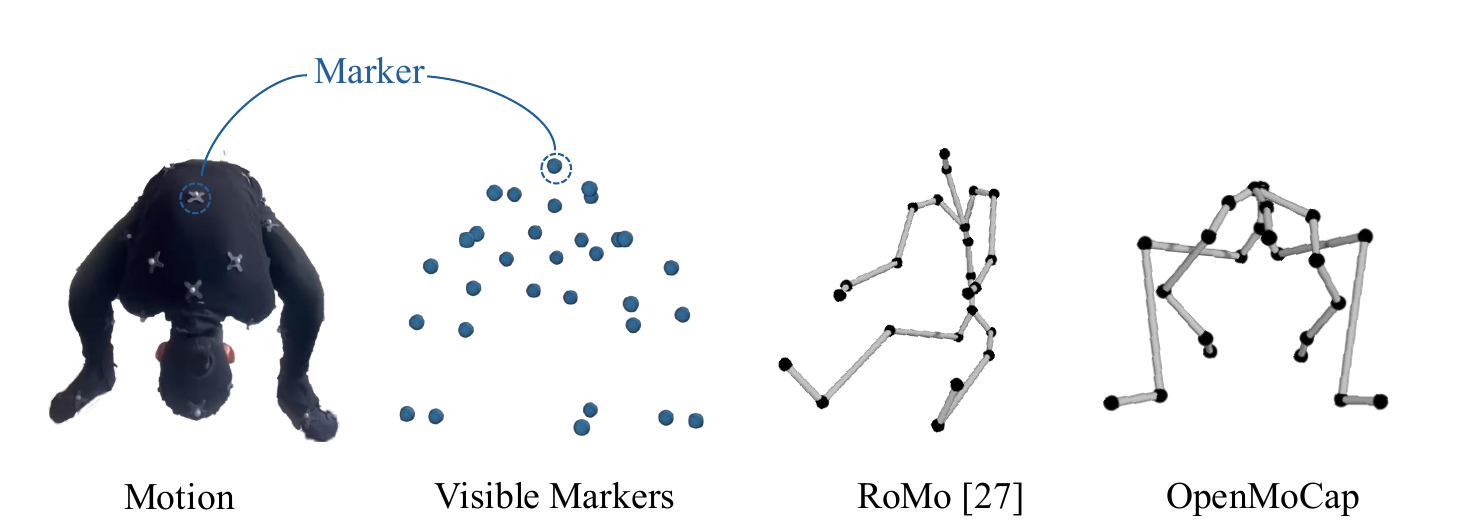}
   \caption{Performance of MoCap in real occlusion scenario. A MoCap actor performing a squatting action with arms extended backward. Significant occlusions occur on markers placed on the abdomen, chest, and forehead. OpenMoCap provides a comparatively reliable solution.}
   \label{fig:intro_performence}
\end{figure}

Albeit inspiring, we observe significant performance degradation when deploying state-of-the-art (SOTA) system \cite{pan2024romo} in real-world production environments due to marker occlusion. Marker occlusion in MoCap occurs when markers are blocked by body parts, obstacles, or environmental structures, which prevents cameras from capturing reflected infrared signals and determining the spatial locations of markers. As illustrated in Fig.\ref{fig:intro_ray}, the markers on the back of the body are visible, whereas those on the abdomen and chest are occluded by the body. Existing method \cite{pan2024romo} fails to reconstruct the human skeleton due to significant occlusions affecting markers placed on regions like the chest as depicted in Fig.\ref{fig:intro_performence}.

\begin{figure}
  \centering
  \begin{subfigure}{0.85\linewidth}
    \includegraphics[width=1\linewidth]{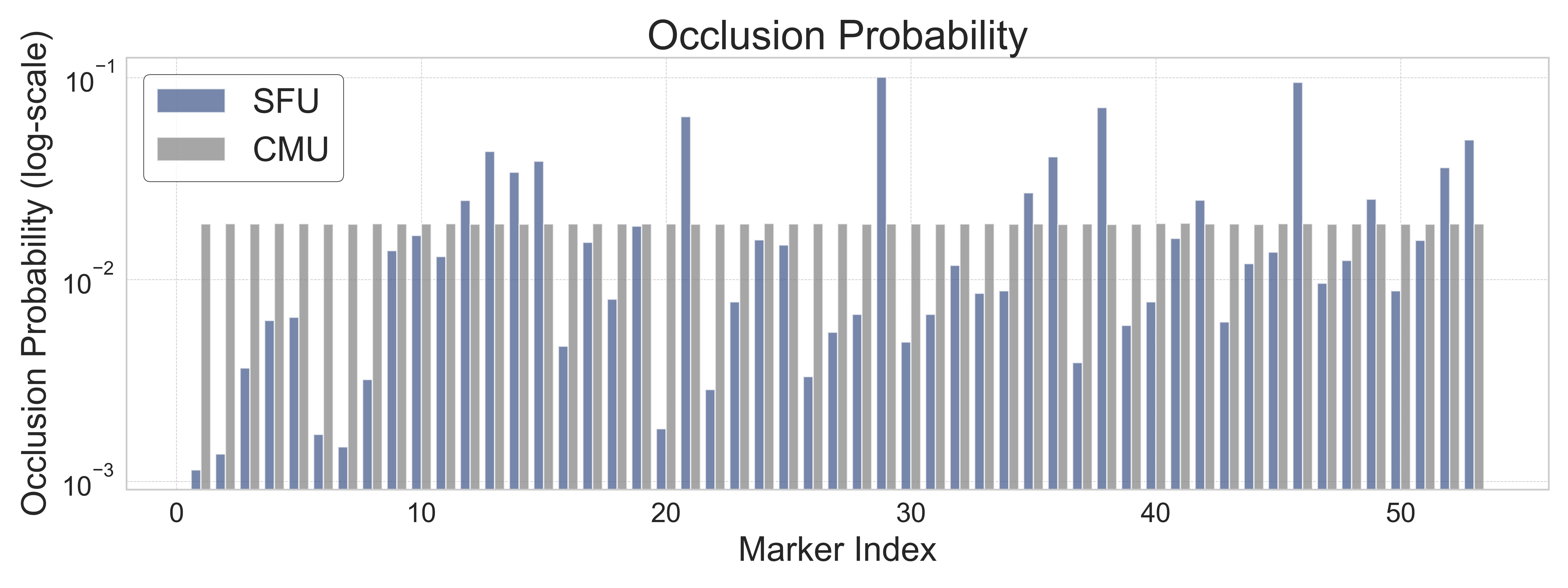}
    \caption{Marker Occlusion Probability.}
    \label{fig:data_occ_prob}
  \end{subfigure}
  \begin{subfigure}{0.85\linewidth}
    \includegraphics[width=1\linewidth]{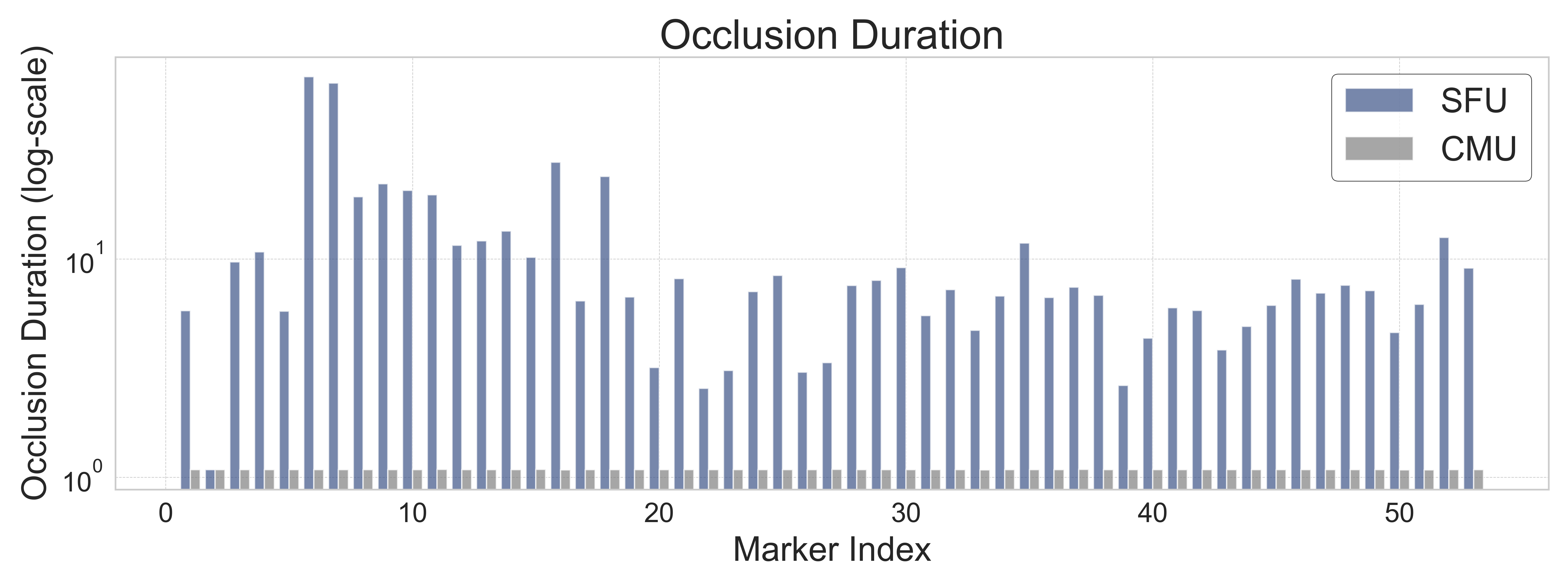}
    \caption{Marker Occlusion Duration.}
    \label{fig:data_occ_dura}
  \end{subfigure}
  \caption{Distribution comparison of marker occlusions in the CMU and SFU MoCap datasets.}
  \label{fig:intro_datacompare}
\end{figure}

Although existing methods exhibit impressive performance on synthetic test sets, they struggle to achieve satisfactory results in solving real MoCap data, like the SFU dataset \cite{sfumocap} . A thorough analysis of this issue has led us to the following two conclusions:

$\bullet$ \textbf{Marker occlusions in real MoCap settings often exhibit high variability and long durations.} Existing datasets typically use random marker occlusions for data augmentation to boost model robustness. We have analyzed occlusion scenarios for markers in both synthetic dataset CMU \cite{cmu}  with random occlusions and the real MoCap dataset SFU as shown in Fig.\ref{fig:intro_datacompare}. Markers are indexed by their proximity relationships. The results reveal that occlusions tend to exhibit certain patterns rather than occurring at random, underscoring a significant mismatch between the occlusion modes in existing synthetic datasets and those observed in real scenarios. This discrepancy becomes problematic as model deployment heavily relies on the assumption that training and testing datasets are independently and identically distributed (i.i.d.). The divergence in data distribution due to marker occlusions leads to a deterioration in model performance, constraining its deployment in real-world settings. This gap highlights the need for more realistic, large-scale training datasets that better reflect real-world marker occlusion patterns.

$\bullet$ \textbf{Existing models lack mechanisms to effectively handle the real characteristics of occlusions.} To address the issue of marker occlusion,  recent initiatives, including LocalMoCap \cite{pan2023locality} and RoMo \cite{pan2024romo}, have introduced methods that utilize visible markers to infer the positions of spatially adjacent occluded markers. However, their underlying assumption does not hold in practice. For example, complex motions often lead to occlusions of multiple adjacent markers simultaneously. This invalid assumption ultimately leads to degraded performance of methods in real-world occlusion scenarios.

To overcome the limitations outlined above, this paper introduces the \textbf{CMU-Occlu dataset}, which better reflects the distribution of marker occlusions in the real world, along with \textbf{OpenMoCap}, a motion solving model designed to address authentic occlusion patterns. Specifically, our work has made targeted innovations in two key dimensions:

\noindent $\bullet$ \textbf{Training Dataset.} We release the CMU-Occlu, a large-scale motion capture dataset that accurately conforms to the characteristics of real marker occlusions. To mitigate the issue of distribution shift between synthetic datasets and real-world data, this study introduces ray tracing algorithms into the generation process of the MoCap dataset for the first time. By simulating various spatial arrangements of infrared cameras in a virtual environment, we have modeled the occlusions caused by obstacles or body parts. This approach significantly enhances the consistency between synthetic data and actual occlusion patterns.

\noindent $\bullet$ \textbf{Solving Mechanism.} This paper proposes a marker-joint chain inference mechanism aimed at accurately reconstructing the positions of both markers and joints. This mechanism incorporates joints and occluded markers as learnable parameters, employing bidirectional chain inference between markers and joints. With the joints serving as intermediaries, this approach establishes long-distance spatial constraints among markers. Furthermore, it enables simultaneous optimization of marker and joint positions, where progressively refined marker estimates contribute to the improved accuracy of joint reconstruction.

To validate the effectiveness of the proposed work, we conducted extensive experiments on both the synthetic MoCap dataset CMU-Occlu and the real MoCap dataset SFU. The experiments demonstrate that CMU-Occlu provides a consistent performance improvement for existing methods \cite{pan2023locality, pan2024romo},  compared to the CMU dataset. Additionally, OpenMoCap surpasses SOTA method, with joint position and joint rotation errors reduced by more than 27\%.

In summary, this paper makes following contributions.
\begin{itemize}
    \item For the first time, a systematic analysis of marker occlusion patterns in real MoCap scenarios is conducted, revealing two fundamental reasons underlying the performance bottlenecks of current SOTA methods: $(i)$ Inconsistencies between the occlusion data distributions in the training sets and the real-world test sets; $(ii)$ The inability of models to effectively establish long-distance spatial constraints.
    \item We introduce the CMU-Occlu dataset, which incorporates realistic marker occlusion patterns and overcomes the limitations of existing optical MoCap datasets with overly simplistic and unrealistic occlusion assumptions. This dataset can serve as a benchmark for evaluation in the field of optical motion capture solving.
    \item We propose a robust MoCap solving model, OpenMoCap, which innovatively implements a marker-joint chain inference mechanism to enhance reconstruction capabilities in scenarios with marker occlusions. Extensive experiments demonstrate that the proposed method surpasses prior state-of-the-art techniques.
    \item Based on the OpenMoCap algorithm, we develop a low-cost MoCap system, MoSen, which eliminates the need for labor-intensive post-processing commonly required in mainstream commercial solutions (e.g., OptiTrack, VICON). By fundamentally transforming the workflow of MoCap repair specialists, our system significantly reduces the overall cost of MoCap and paves the way for its broader adoption.
\end{itemize}

\section{Related Work}
\label{sec:related_work}

\subsection{Motion Data Synthesis}

The CMU Motion Capture (MoCap) dataset~\cite{cmu} is a widely used benchmark in human motion analysis, offering over 2000 high-quality sequences captured via a Vicon system, covering diverse actions like walking, jumping, and dancing. It has served as a foundation for developing and evaluating data-driven MoCap algorithms. SMPL~\cite{loper2023smpl} enables accurate motion reconstruction by parameterizing body shape and pose. AMASS~\cite{cmu, mandery2015kit, muller2009efficient, muller2007mocap} unifies data from multiple MoCap datasets and refines surface representations to provide high-quality motion data.


To improve model robustness, synthetic datasets often apply frame-wise random occlusion~\cite{chen2021mocap, holden2018robust} or long-term occlusion of a single marker~\cite{pan2023locality}. In real-world settings, however, occlusions are typically sustained and affect multiple neighboring markers, creating a distribution gap that limits model applicability in practical MoCap scenarios.

\subsection{Motion Capture Solving}
Recently, MoCap has become a research hotspot. UUO-Mocap~\cite{milef2024towards} tackles motion capture in unstructured, unlabeled video settings by leveraging body priors and handling partial-body observations. SportsCap~\cite{chen2021sportscap} addresses challenging sports scenarios with a framework that jointly captures 3D motion and fine-grained actions using structured priors and a multi-stream spatio-temporal GCN.

Optical motion capture, the most precise and widely adopted method, estimates body pose from marker trajectories. Traditional marker-based solving methods~\cite{aristidou2018self, feng2014exploiting, liu2014automatic, aristidou2013real} rely on geometric constraints and optimization algorithms under specific assumptions. Recently, deep learning has driven advances in marker tracking and data reconstruction. MoSh++~\cite{mahmood2019amass} estimates body pose via frame-wise parameter optimization. MoCap-Solver~\cite{chen2021mocap} encodes skeletal structure, marker layout, and motion separately, then jointly decodes them. LocalMoCap~\cite{pan2023locality} exploits local marker dependencies, completing occluded positions from neighbors and applying GCNs for motion inference. Damo~\cite{kim2024damo} improves generalization across marker layouts, while RoMo~\cite{pan2024romo} addresses marker mislabeling and positional noise. Although data-driven methods are more robust, most are not explicitly designed for real-world occlusion, leading to performance degradation when assumptions about clean inputs are violated.


In other fields~\cite{chen2022structural, bao2022fusepose}, methods like MAE~\cite{he2022masked} enhance learning by masking parts of the input and training on the visible data. OpenMoCap adopts this idea to recover occluded information.

\section{CMU-Occlu Dataset Synthesis}
\label{sec:dataset}

Marker occlusion is inevitable in optical motion capture. Existing synthetic datasets rely solely on random occlusion methods for data augmentation. However, this approach does not accurately reflect occlusion patterns observed in real-world motion capture scenarios, thus limiting the effectiveness and efficiency of current pre-trained models when deployed in production environments.

\subsection{Preliminary: Marker Capture}
Here, we briefly describe the working principle of optical motion capture systems. In such systems, calculating the three-dimensional coordinates of markers relies on satisfying a co-visibility constraint: spatial position of one marker can be reconstructed using epipolar geometry \cite{hartley2003multiple} only if it is simultaneously captured by at least two infrared cameras. As illustrated in Fig.\ref{fig:intro_ray}, two stationary infrared cameras are positioned at the top, and the blue spheres attached to the body represent markers. In this example, when the subject performs a forward-bending movement:

\noindent $\bullet$ \textbf{Visible Marker}: The marker on the back remains visible to both cameras, satisfying the co-visibility condition, allowing its accurate 3D coordinates to be calculated through triangulation.

\noindent $\bullet$ \textbf{Occluded Marker}: Markers on the abdomen and chest become obscured by body, resulting in fewer than two cameras capturing their reflected signals. Consequently, the system cannot form valid observation equations for their reconstruction, and these markers are classified as occluded.

Considering the working principles of MoCap systems, marker occlusion depends on various factors, including the number of cameras, their spatial arrangement, and the complexity of movements performed. Consequently, the resulting occlusion patterns often involve extensive occlusions and prolonged occlusion periods for individual markers.

\subsection{Dataset Construction}
To incorporate realistic marker occlusion patterns into optical motion capture datasets, we propose an improved version of the original CMU dataset \cite{cmu}, termed CMU-Occlu. This dataset leverages a parallel implementation of the Möller–Trumbore algorithm \cite{moller1997fast}, integrating real MoCap scenario configurations. As a result, CMU-Occlu effectively bridges the gap between existing synthetic datasets and real-world occlusion patterns.

Specifically, an infrared ray is defined by its origin \( \mathbf{r}_0 \) and direction vector \( \mathbf{d} \):
\begin{equation}
    \mathbf{R}(t) = \mathbf{r}_0 + t\mathbf{d}.
\end{equation}
At the same time, the triangular mesh of the human body is defined by its vertices \( \mathbf{v}_1, \mathbf{v}_2, \mathbf{v}_3 \), and the marker \( \mathbf{p} \) located on a mesh triangle can be expressed as:
\begin{equation}
    \mathbf{p} = (1 - \beta - \gamma) \mathbf{v}_1 + \beta \mathbf{v}_2 + \gamma \mathbf{v}_3.
\end{equation}
To determine whether the human body occludes the infrared ray from capturing a marker, it is necessary to check whether the mesh intersects with the ray, and whether the intersection point lies closer to the camera than the marker itself.
\begin{equation}
    \mathbf{r}_0 + t\mathbf{d} = (1 - \beta - \gamma) \mathbf{v}_1 + \beta \mathbf{v}_2 + \gamma \mathbf{v}_3,
\end{equation}
\begin{equation}
    \mathbf{is\_occluded} = (t_{\text{closest}} < \| \mathbf{p} - \mathbf{r}_0 \|),
\end{equation}
where \( t_{\text{closest}} \) is the value of \( t \) for the nearest valid intersection point.

\begin{figure}
  \centering
  \begin{subfigure}{0.48\linewidth}
  \includegraphics[width=1\linewidth]{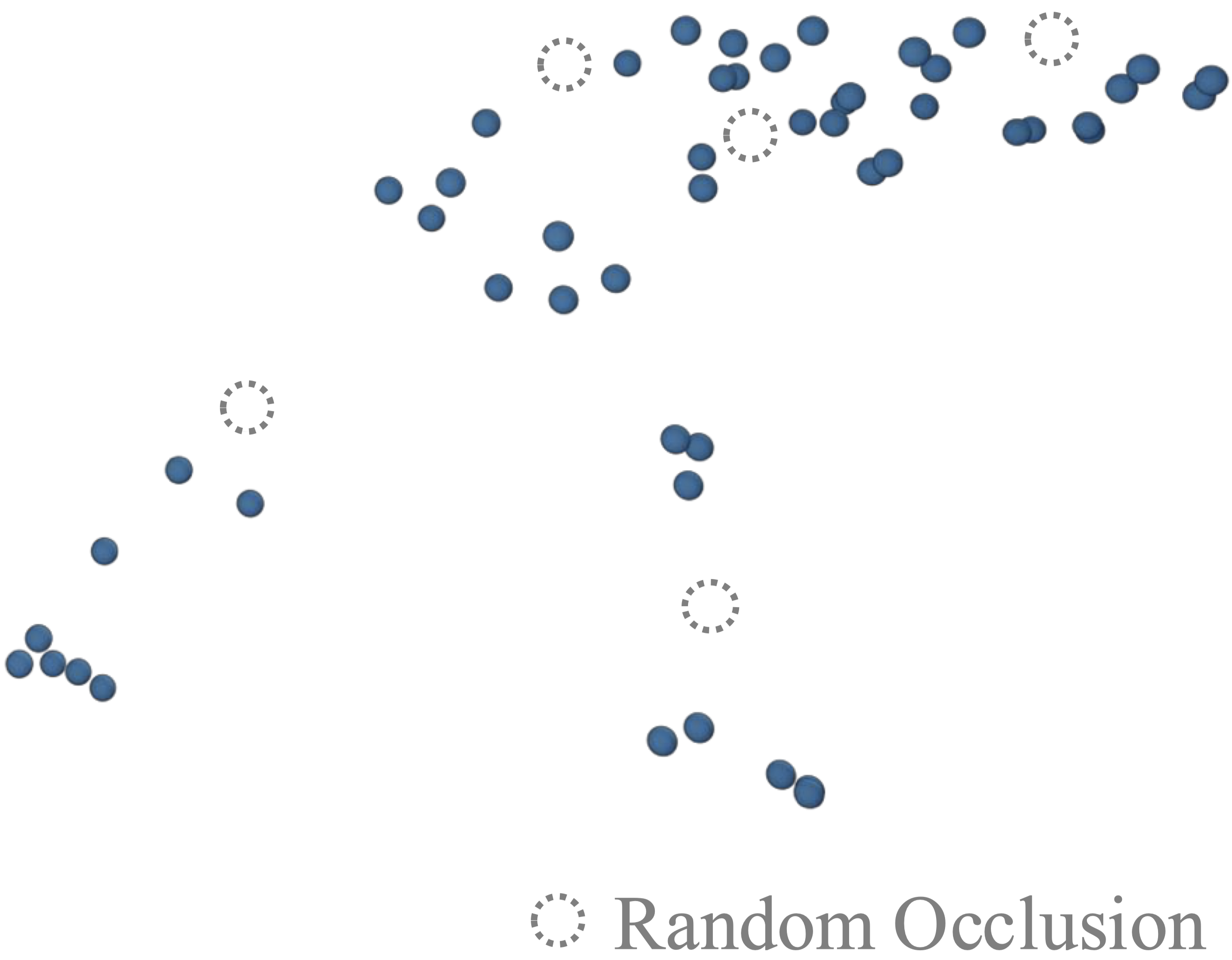}
    \caption{CMU Dataset.}
    \label{fig:frame_cmu}
  \end{subfigure}
  \hfill
  \begin{subfigure}{0.48\linewidth}
    \includegraphics[width=1\linewidth]{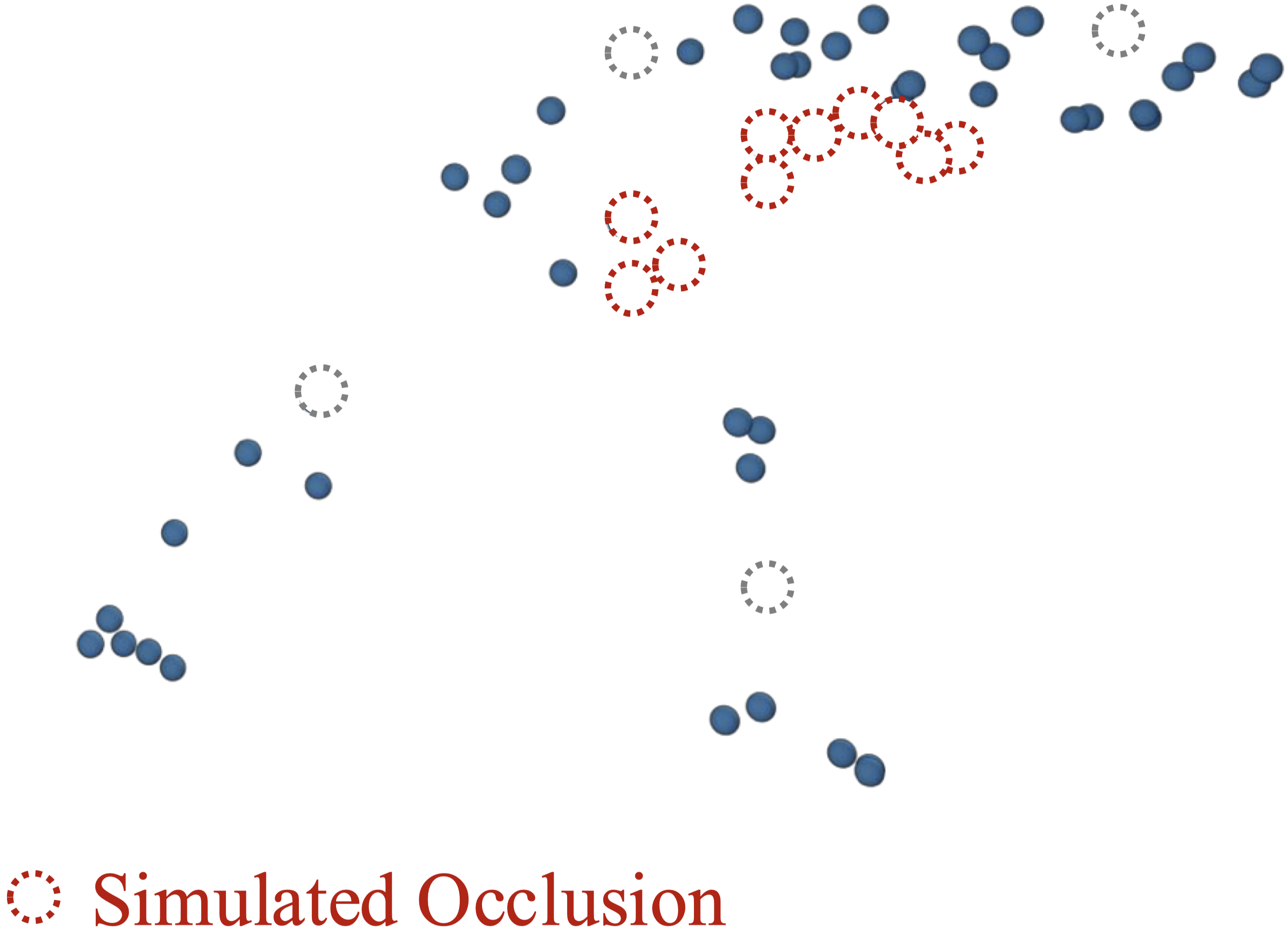}
    \caption{CMU-Occlu Dataset.}
    \label{fig:frame_occ}
  \end{subfigure}
  \caption{Comparison of marker occlusion patterns of different datasets.}
  \label{fig:frame_compare}
\end{figure}

Through this approach, we incorporate simulated marker occlusion into the CMU-Occlu dataset. As illustrated in Fig.\ref{fig:frame_compare}, the corresponding frames of markers from both the CMU dataset and the CMU-Occlu dataset are visualized.

\begin{figure}
  \centering
  \begin{subfigure}{0.48\linewidth}
  \includegraphics[width=1\linewidth]{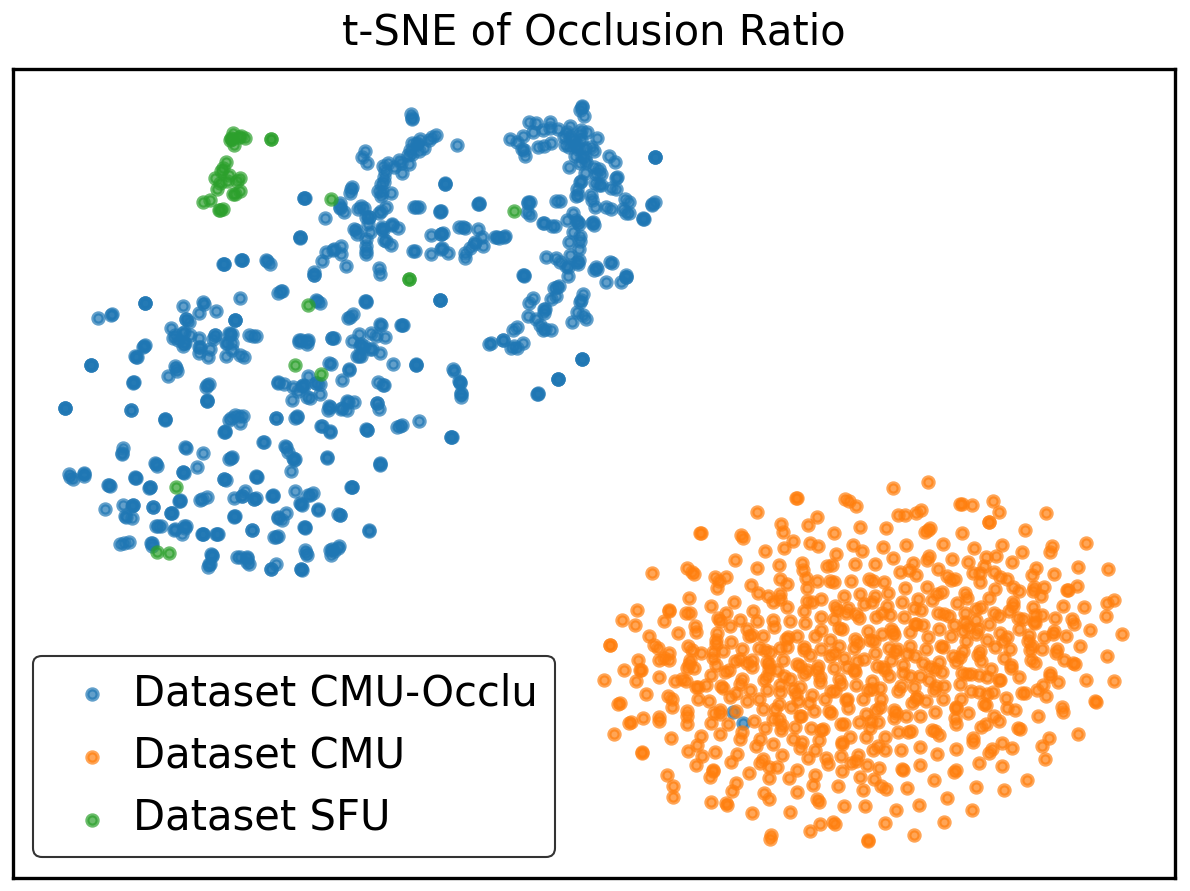}
    \caption{Marker Occlusion Ratio.}
    \label{fig:tsne_prob}
  \end{subfigure}
  \hfill
  \begin{subfigure}{0.48\linewidth}
    \includegraphics[width=1\linewidth]{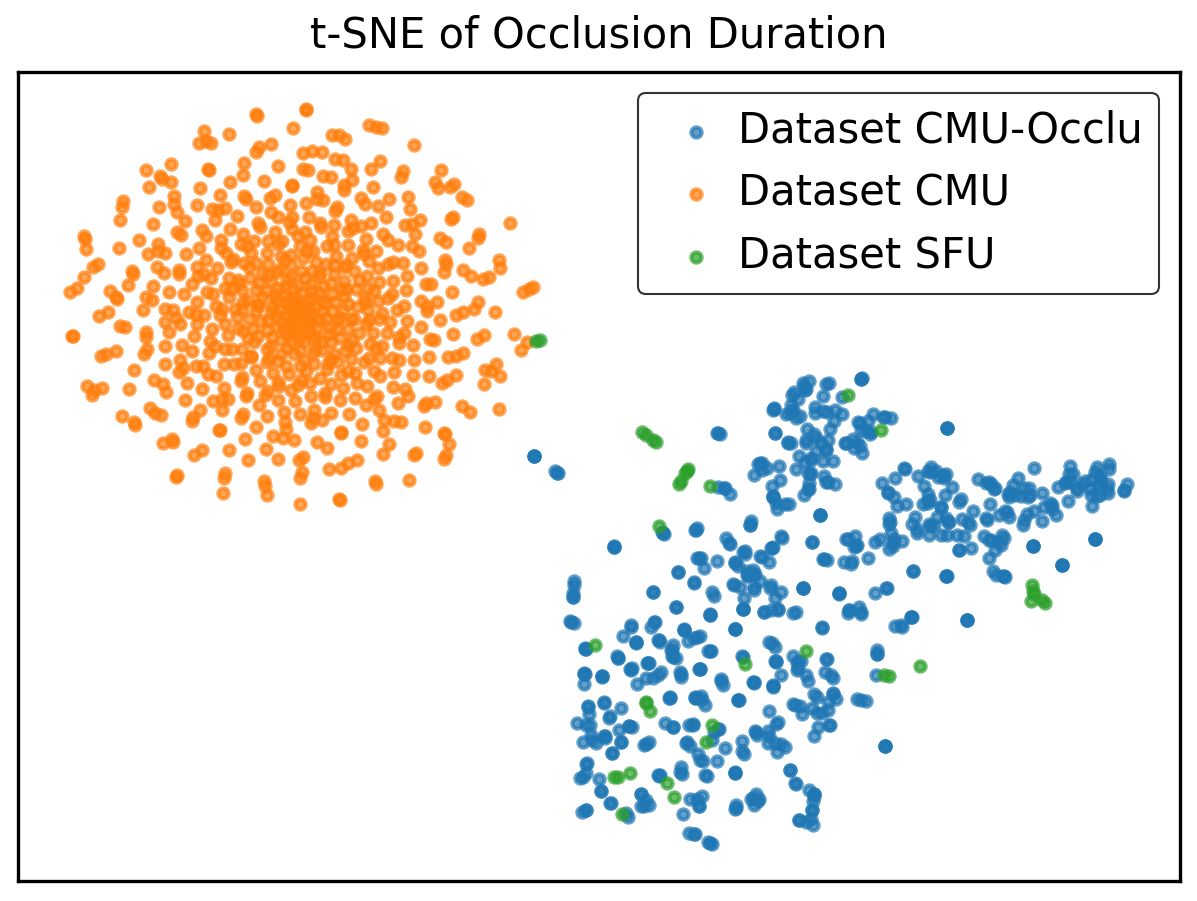}
    \caption{Marker Occlusion Duration.}
    \label{fig:tsne_dura}
  \end{subfigure}
  \caption{Comparison of marker occlusion distribution characteristics across datasets.}
  \label{fig:dataset}
\end{figure}

To facilitate training more robust models, the proposed CMU-Occlu dataset encompasses both random and simulated occlusion patterns. Specifically, the simulated occlusion patterns are generated by selecting four cameras through comparing the Kullback–Leibler (KL) divergence of occlusion distributions with the real-world SFU dataset. Additionally, the four camera layouts with the highest divergence are chosen, and oversampling \cite{buda2018systematic} is employed to address the imbalance problem arising from an insufficient number of occluded MoCap frames.

As shown in Fig.\ref{fig:dataset}, we conduct a comparative analysis of marker occlusion distributions across the CMU, CMU-Occlu, and the real-world MoCap dataset SFU \cite{sfumocap}. The comparison encompasses both the occlusion probability of individual markers and the duration of occlusion for each marker. Results show that the occlusion distribution of CMU-Occlu aligns more closely with that of SFU, whereas CMU exhibits a distinctly different pattern.

To ensure versatility and ease of use across various research and production environments, we will publicly release our dataset generation method as open-source, along with interfaces supporting different parameter configurations to accommodate diverse application scenarios.
\section{Architecture}

\begin{figure*}[t]
  \centering
   \includegraphics[width=1\linewidth]{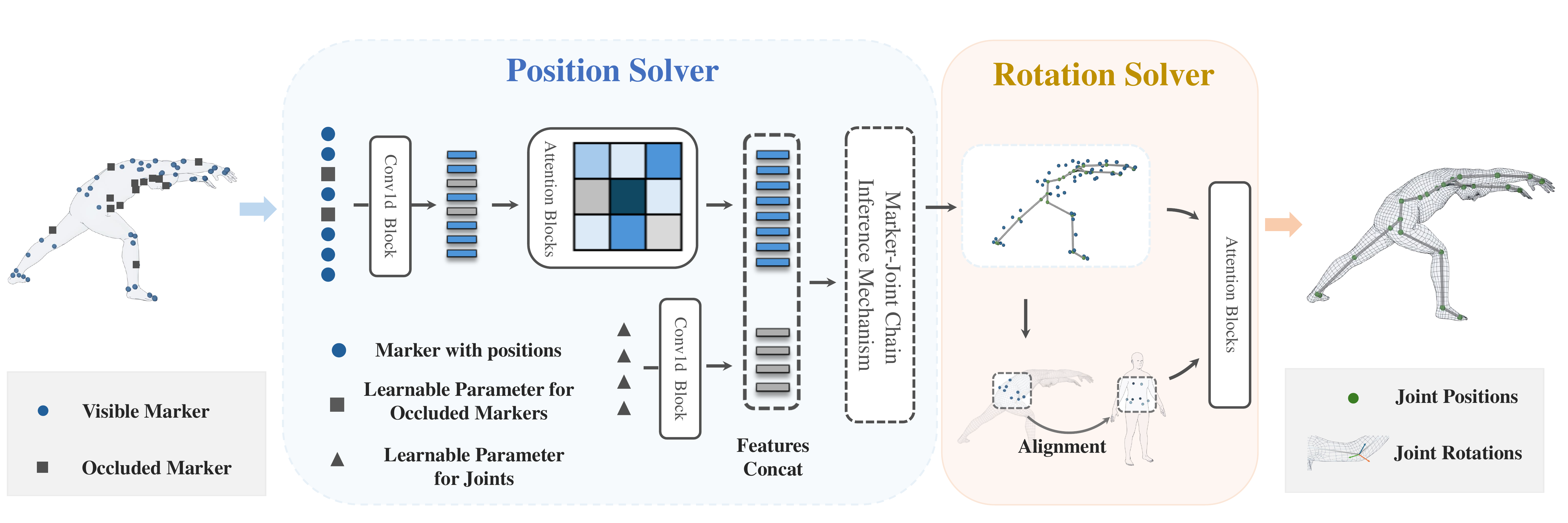}

   \caption{Overview of the proposed multi-stage framework, OpenMoCap. The framework takes raw markers with occlusions as input. A Position Solver first estimates the positions of all markers and joints through the Marker-Joint Chain Inference Mechanism. These positions are then used to align the input for a Rotation Solver, which predicts joint rotations using a stacked attention-based architecture.}
   \label{fig:archi}
\end{figure*}

Given a captured point cloud of visible markers in a single frame, our goal is to accurately estimate joint positions and rotations, while simultaneously reconstructing the positions of markers that may be occluded or displaced. To address this challenge, we propose a multi-stage framework that decouples position estimation from rotation estimation.



\subsection{Decoupled MoCap Architecture}
\label{sec:decoupled}
Different joints play distinct roles in human motion, with waist-region joints (e.g., pelvis) being critical for determining global pose. To ensure accurate modeling, most methods assume partial visibility of key markers. For instance, to aid model convergence, training data is typically aligned to a standard pose using eight waist markers. This alignment introduces two key limitations:

\noindent $\bullet$ \textbf{Occlusion Sensitivity}. Alignment methods such as Singular Value Decomposition (SVD) require at least three pairs of corresponding visible points between two point clouds. Consequently, if one of key markers remain continuously occluded, models like MoCap-Solver\cite{chen2021mocap} cannot function properly.

\noindent $\bullet$ \textbf{Irreversible Error Propagation}. Directly performing spatial alignment using partially missing critical markers \cite{pan2023locality,pan2024romo} introduces alignment errors at the preprocessing stage. These initial errors propagate iteratively through the global skeletal model, ultimately causing substantial distortions in the reconstructed motion.

To address critical data loss caused by marker occlusions, we propose a decoupled MoCap solving architecture that postpones the spatial alignment process. This architectural design stems from a fundamental insight: solving for joint and marker positions constitutes a linear problem that does not heavily depend on spatial alignment outcomes. Conversely, joint rotation estimation is inherently more challenging, characterized as a nonlinear problem that benefits significantly from spatial alignment to facilitate training convergence.

Following this design principle, the overall architecture of our approach, OpenMoCap, is illustrated in Fig.\ref{fig:archi}. The architecture consists of two primary components: a Position Solver and a Rotation Solver. The Position Solver takes the raw marker positions as input to compute joint positions. It also reconstructs positions of all markers, filling in occluded markers and correcting positions of displaced markers. Subsequently, critical reference markers are spatially aligned with their corresponding markers in a T-pose configuration to perform a transformation. The whole aligned markers are then used in the Rotation Solver, which computes the joint rotations of the human body.

\subsubsection{Position Solver}



To handle varying levels of marker occlusion, we represent occluded markers as shared learnable parameters. This allows the network to learn their feature distributions, avoiding distortion from fixed placeholders like zeros. Convolutional and attention modules extract features from visible markers, which are then propagated to occluded ones.

Joint position solving is formulated as a generative task, where the input includes extracted features and $N$ shared parameters that guide the generation of joint representations. The decoder, based on the Marker–Joint Chain Inference Mechanism, computes attention distributions. The loss function for position prediction is defined as follows:
\begin{equation}
  L_{P} = \lambda_{1} L_{M_{occ}} + \lambda_{2} L_{M_{shift}} + \lambda_{3}  L_{J}.
  \label{eq:lossp}
\end{equation}

Since the number of occluded markers is relatively small compared to the total number of markers, we emphasize the importance of marker prediction by separating the marker completions and marker refinements into two distinct losses. $ L_{M_{occ}}$ represents the error in solving the positions of the occluded markers. $L_{M_{shift}}$ represents the error in correcting the positions of the shifted markers, and $L_{J}$ represents the error in solving the positions of the human joints. All three errors are computed using Euclidean distance. The parameters $\lambda_{1}$, $\lambda_{2}$ and $\lambda_{3}$ will be discussed in the experimental section.
\vspace{-5pt}

\subsubsection{Rotation Solver}
After processing through the Position Solver, we obtain accurate joint and marker positions. Leveraging this precise and detailed input, we construct the Rotation Solver to estimate joint rotations. 

As noted earlier, the positions of key reference markers are used to eliminate global transformations. Considering the forward kinematics of the human body, joint rotations can be inferred by fitting the skeletal model to the observed marker, indicating a strong correlation between marker locations and joint rotations. To capture this relationship, the rotation solver incorporates stacked attention modules. 

In terms of loss function design, instead of using hierarchical weighting, we compute rotation errors for each joint independently. This design offers two key advantages:
(i) it ensures equal weighting of rotation loss across all joints, thereby preventing error accumulation and propagation; (ii) hierarchical schemes impose fixed parent–child dependencies in joint rotations. We relax these constraints to allow more flexible motion modeling.

Finally, to ensure continuity during rotation regression training, we adopt the 6D representation \cite{zhou2019continuity} for computing rotation errors. Notably, the network operates without relying on temporal correlations, reducing preprocessing overhead and enabling real-time motion MoCap with deep learning.

\subsection{Marker-Joint Chain Inference Mechanism}
\label{sec:chain}

In real-world motion capture environments, occlusion of a marker by the body or external obstacles often leads to the simultaneous occlusion of neighboring markers. To address the challenges posed by such occlusion patterns, we conduct an in-depth investigation into the relationship between markers and joints, and accordingly design a marker-joint chain inference mechanism. This mechanism is driven by the key insight that markers and joints are mutually constrained. 


\textbf{Bidirectional Inference} represents the mutual relationship between the positions of markers and joints. Joint features extracted from imperfect marker positions help integrate contextual information, thereby contributing to more accurate marker position refinement.

\begin{figure}[t]
  \centering
   \includegraphics[width=1\linewidth]{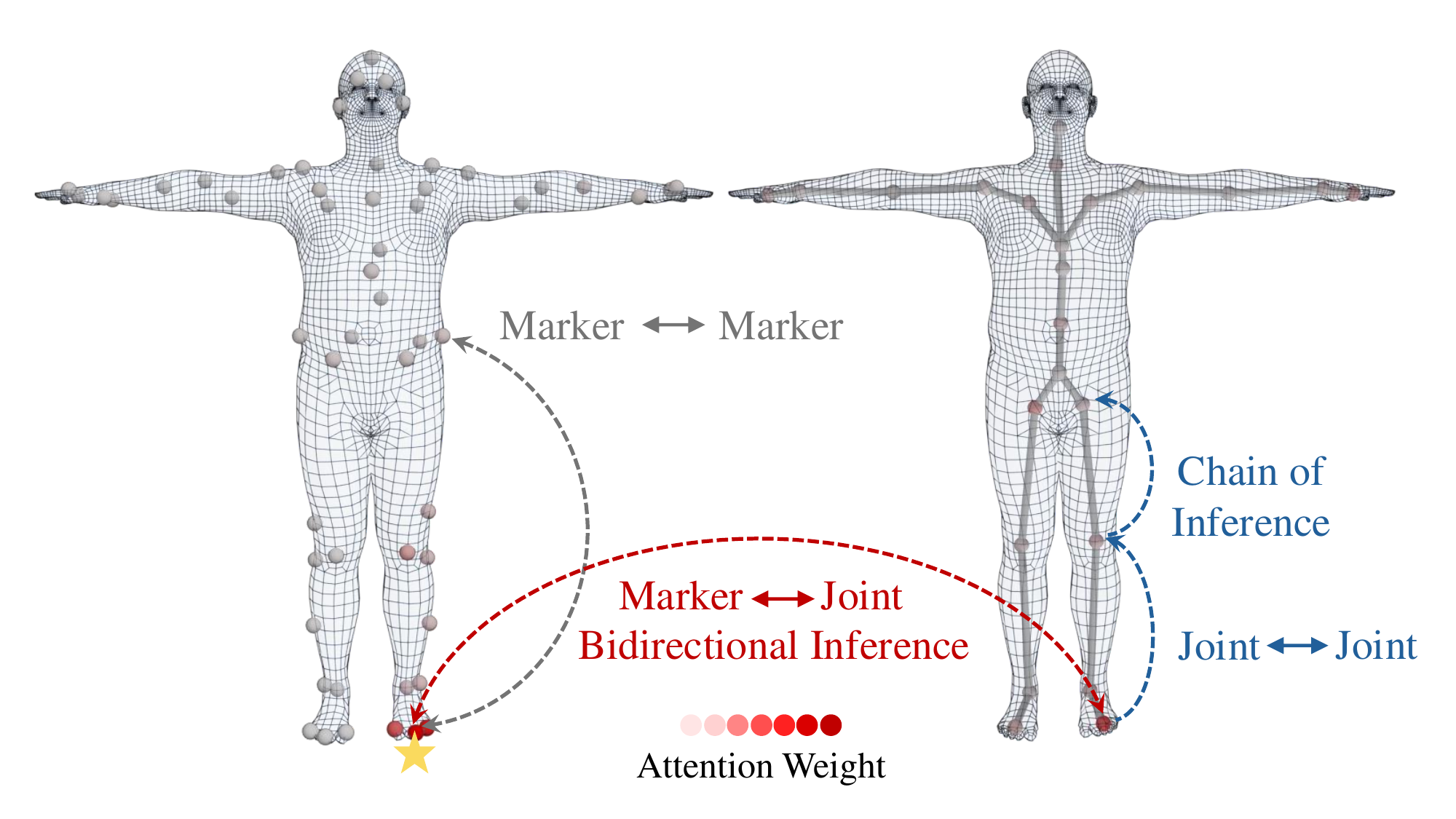}

   \caption{Marker-Joint Chain Inference Mechanism. The figure highlights a marker on the front of the foot (indicated by a star) and visualizes attention weights across other markers and joints, with red intensity indicating attention strength. Long-range dependencies are captured through bidirectional marker–joint reasoning and joint chain inference.}
   \label{fig:chain}
\end{figure}

This insight is validated in our experiments. As illustrated in Fig.\ref{fig:chain}, the red color gradient indicates the weights from one attention layer, with lighter shades corresponding to lower attention values. The yellow pentagon marks the target marker of interest. When estimating its position, the model relies not only on information from other markers but also on that from the joints. By learning the bidirectional correlations between markers and joints, the network iteratively integrates information from both sources, enabling mutual refinement and unified optimization of the final result.

\textbf{Chain of Inference} represents the formation of information pathways. As shown in Fig.\ref{fig:chain}, a marker exhibits limited dependency on distant markers. This sheds light on the limitations of previous approaches that focus exclusively on mutual completion among markers, as they fail to effectively leverage all available information for accurate completion. The marker on the right toe is closely associated with the corresponding toe joint and even establishes a connection to the left hip joint through inter-joint relationships. When the foot markers are heavily occluded, the markers near the hip can help constrain the possible positions of the foot markers by predicting joint positions and performing chain-like reasoning. As a result, long-range dependencies are successfully established.


Previous methods typically infer joint positions and rotations from initially corrected marker positions. In contrast, the Marker-joint chain inference mechanism introduces a key distinction: it treats joints as intermediate nodes to establish long-range dependencies among markers. Specifically, marker information can be propagated through the close connections between nearby joints and other related joints. Since joint positions are tightly coupled with marker positions, they can in turn help refine or complete missing or distant marker observations. This simultaneous optimization of markers and joints leads to more accurate reconstruction results. In particular, precise and complete estimation of markers around the waist is crucial for subsequent spatial alignment procedures.

We formalize the modeling process as follows. The entire mechanism can be interpreted as an information diffusion process over a weighted directed graph. Specifically, let $M$ denote the set of all markers, $J$ the set of joints to be predicted, and $V$ the complete set of nodes, where:
\begin{equation}
    M = \left \{ m_1,...,m_M \right \}, J=\left \{ j_1,...,j_J \right \} , V = M \cup J.
\end{equation}
We define the initial state of the directed graph as $h^{(0)}$, and the weighted propagation matrix at step $t$ as $P^{(t)}$. Accordingly, the representation of the $k$-th marker after $L$ steps can be expressed as:
\begin{equation}
    h_k^{(L)} = \sum_{v=1}^{M+J}\left [ P^{(L-1)}P^{(L-2)}\cdots P^{(0)} \right ]_{k,v}h_v^{(0)}.
\end{equation}
Similarly, the state of the $k$-th joint after $L$ steps can be computed as:
\begin{equation}
    h_{M+k}^{(L)} = \sum_{v=1}^{M+J}\left [ P^{(L-1)}P^{(L-2)}\cdots P^{(0)} \right ]_{M+k,v}h_v^{(0)}.
\end{equation}
As shown in Fig.\ref{fig:chain}, the chain reasoning between different joint positions across multiple steps can thus be expressed as:
\begin{equation}
    [P^{(s)}P^{(s-1)}]_{j_c, j_a} = \sum_{j_b}P_{j_c, j_b}^{(s)}P_{j_b, j_a}^{(s-1)}.
\end{equation}
\section{Experiments}
\label{sec:experiments}

\subsection{Experimental Details}
\label{sec:details}

\subsubsection{Parameter Settings and Training Environment}
In our experiment, the input markers are centralized by subtracting the centroid position. The occluded markers and input joints are set to different shared parameters respectively.

As for the weight of loss in position solving network, we use
\begin{equation}
 L_{P} = L_{M_{occ}} + L_{M_{shift}} + 2 * L_{J}.
  \label{eq:lossps}
\end{equation}
Since accurate positions of markers are very important for the operation of  rotation network, the position solving loss of marker is set to the same weight as the joint position.

The whole network is trained on 1 GeForce RTX 4090 with 24GB memory, and the batch size is set to 256.

\subsubsection{Dataset}
A dataset of superior quality, encompassing a wide variety of motion types and body sizes, is crucial for enhancing the generalization capabilities of models. Three datasets are used in our experiments. In addition, we collected a set of real-world MoCap sequences to qualitatively evaluate and compare the performance of different approaches.

The first dataset is generated by driving the SMPL model using pose parameters from the CMU MoCap dataset \cite{cmu} and shape parameters from the CAESAR dataset \cite{robinette2002civilian}. We refer to this dataset as the CMU dataset. We apply the corruption function proposed by Holden \cite{holden2018robust}. The dataset consists of 5k synthetic MoCap sequences, totaling 8m frames, with 1,700 characters and 5,100 marker configurations. As described in Sec.\ref{sec:dataset}, the second dataset we use is the CMU-Occlu Dataset. The third dataset is SFU Motion Capture Database \cite{sfumocap}. This real MoCap dataset includes motion capture recordings from 8 actors, with a total of 44 manually annotated sequences.


\subsubsection{Evaluation Metrics}
Given the positions of markers as input, the model outputs the positions and rotations of joints. In the following experiments, we use Joint Position Error (JPE) to represent the Euclidean distance between the predicted results and ground truth (GT). We also use Joint Orientation Error (JOE) to measure the discrepancy between the predicted and actual joint rotation angles.

\begin{table}[t]
  \centering
  \renewcommand{\arraystretch}{1.2}
  \resizebox{1\linewidth}{!}{
  \begin{tabular}{ccccccc}
   \toprule
   \multirow{2}{*}{}& \multirow{2}{*}{}& \multirow{2}{*}{MoSh++} & \multirow{2}{*}{\parbox{1cm}{\centering MoCap- \\ Solver}} & \multirow{2}{*}{\parbox{1cm}{\centering Local \\ MoCap}}& \multirow{2}{*}{RoMo} & \multirow{2}{*}{\parbox{1cm}{\centering Open \\ MoCap}} \\ 
   &&&&&&
    \\
    \midrule
   \multirow{2}{*}{\textbf{CMU}} 
       & JPE (cm)  & 2.58  & 2.56  & 0.94 & 0.89   & \textbf{0.41}\\
            & JOE ($^\circ$)  & 9.40  & 6.51 & 3.59 &  3.43   & \textbf{2.52}\\
     \cmidrule{1-7}
            \multirow{2}{*}{\parbox{1cm}{\centering \textbf{CMU- \\ Occlu}}} 
            & JPE (cm)  & 2.72 & 2.95&1.23 & 1.16&   \textbf{0.46}\\
            & JOE ($^\circ$)  & 9.68 & 6.83 &3.80 &3.54 & \textbf{2.60}\\
   \bottomrule
  \end{tabular}
  }
  \vspace{2pt}
\caption{Comparison with other methods on different datasets.}
\label{tab:works_compare}
\end{table}

\begin{figure}
  \centering
  \begin{subfigure}{0.48\linewidth}
  \includegraphics[width=1\linewidth]{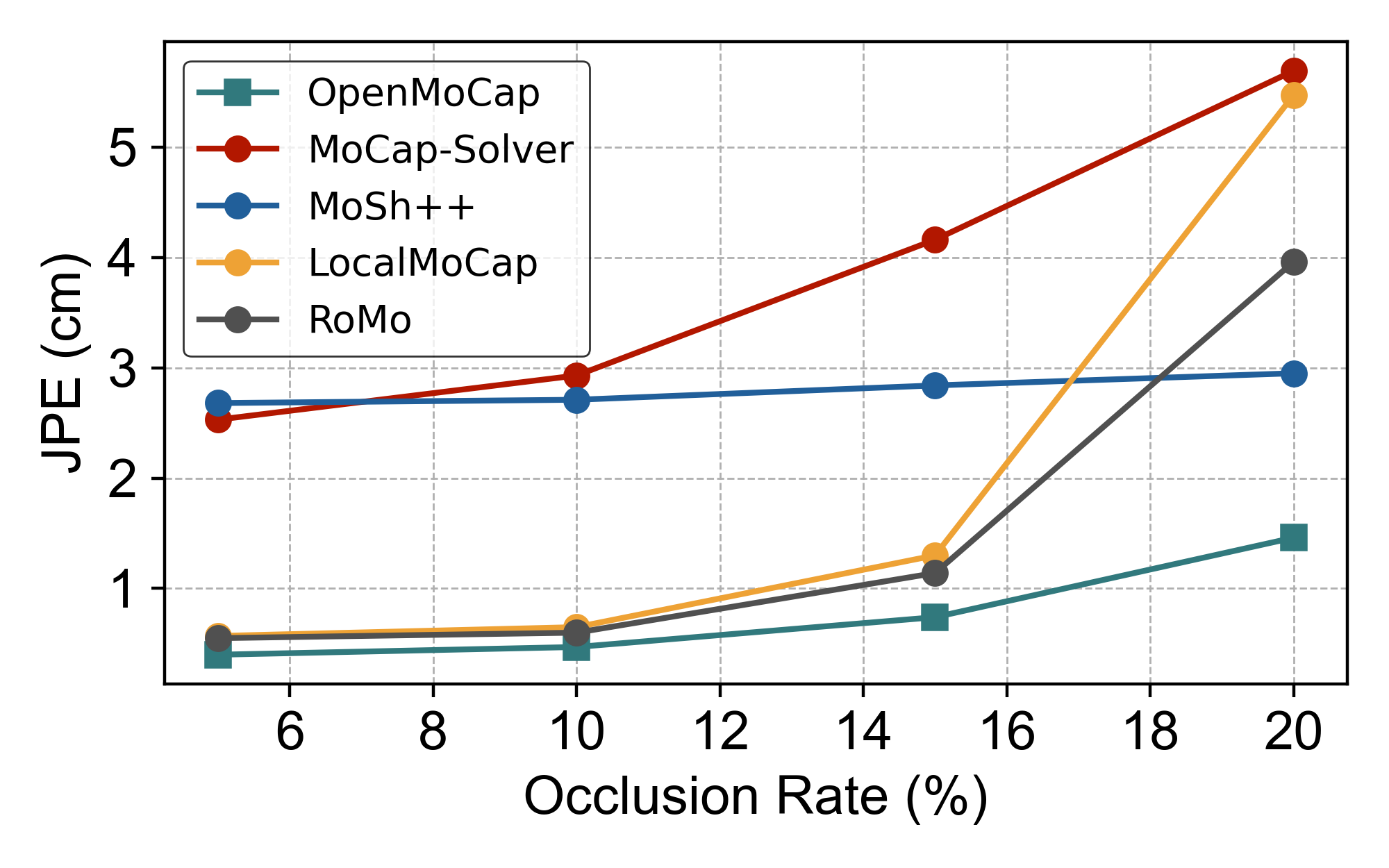}
    \caption{Joint Position Error (JPE).}
    \label{fig:occrate_jpe}
  \end{subfigure}
  \hfill
  \begin{subfigure}{0.48\linewidth}
    \includegraphics[width=1\linewidth]{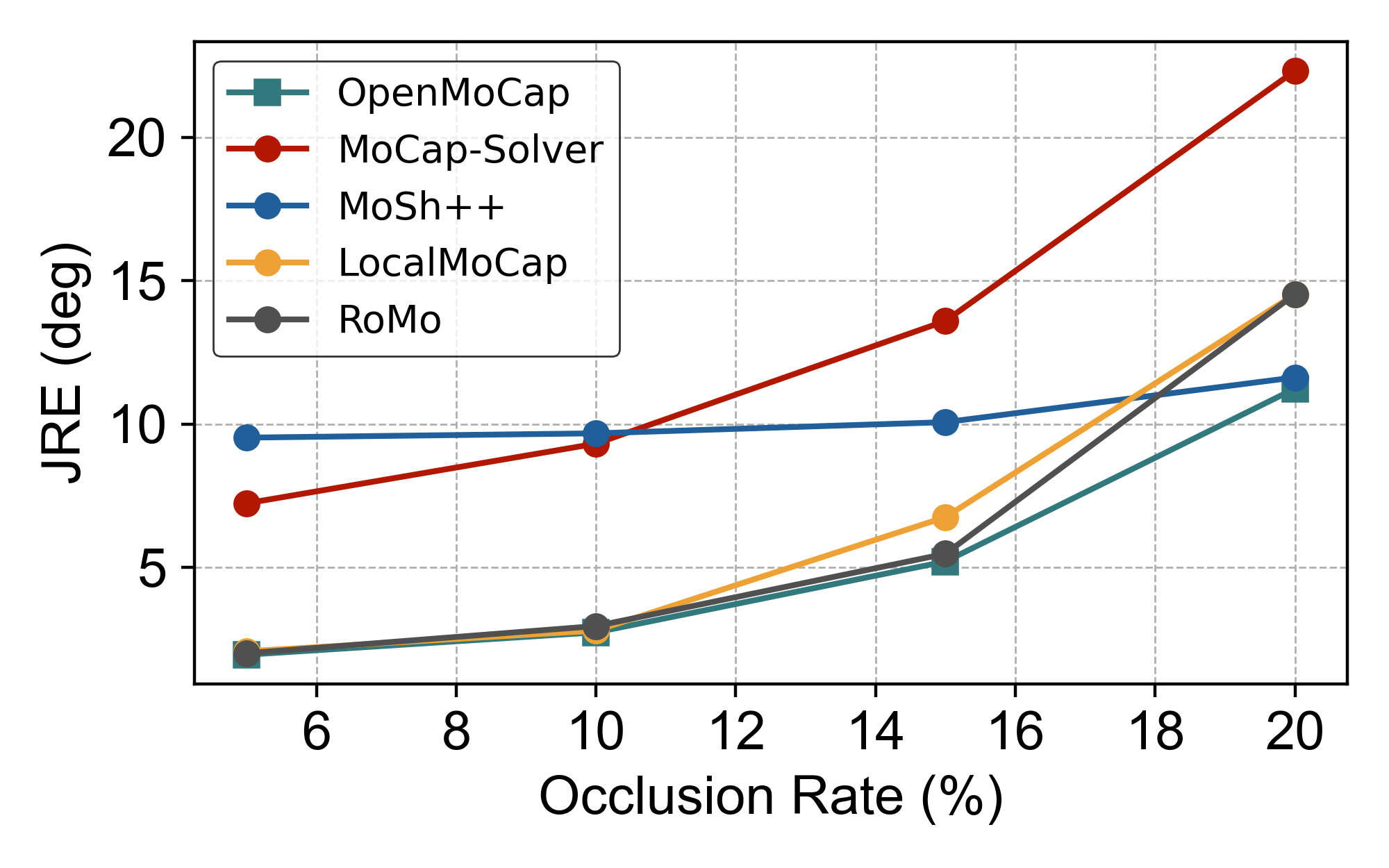}
    \caption{Joint Rotation Error (JRE).}
    \label{fig:occrate_jre}
  \end{subfigure}
  \caption{Error variation of different methods under varying marker occlusion ratios.}
  \label{fig:exp_occrate}
\end{figure}

\subsection{Approach Comparisons}
\label{sec:approach_compare}
Each method is independently trained and tested on the CMU and CMU-Occlu datasets, respectively. Both datasets contain marker occlusions and positional perturbations. As shown in Tab.\ref{tab:works_compare}, the increased level of occlusion in CMU-Occlu leads to a slight performance degradation across all methods.

MoSh++~\cite{mahmood2019amass}, as a parameter optimization method, relies heavily on temporal continuity and struggles with frame-wise corrupted data. MoCap-Solver~\cite{chen2021mocap} depends on consistently visible key markers, limiting its robustness in real-world settings. LocalMoCap~\cite{pan2023locality} and its successor RoMo~\cite{pan2024romo} address occlusions by interpolating missing markers and predicting their positions using neural networks. This approach achieves favorable results on the CMU dataset with randomly simulated occlusions, where occlusion durations are short and adjacent markers can compensate for each other. Among these methods, OpenMoCap achieves the best performance on both the CMU and CMU-Occlu datasets.

To further evaluate the robustness of different methods under varying levels of occlusion, we divide the CMU-Occlu test set into four subsets based on occlusion severity: 5\%, 10\%, 15\%, and 20\%. For a given sequence, if more than half of its frames contain over 20\% of occluded markers, it is assigned to the 20\% occlusion group. Models trained on the CMU-Occlu dataset are then tested under each occlusion condition, and the results are shown in Fig.\ref{fig:exp_occrate}.

\begin{table}[t]
  \centering
  \renewcommand{\arraystretch}{1.2}
  \resizebox{0.9\linewidth}{!}{
  \begin{tabular}{ccccc}
   \toprule
   & & \multirow{2}{*}{\parbox{1.5cm}{\small \centering w/o marker-\\joint chain}} & \multirow{2}{*}{\parbox{1.7cm}{\small \centering w/o decoupled \\ architecture}}&  \multirow{2}{*}{\parbox{1cm}{\small \centering Open \\ MoCap}} \\ 
   &&&&
    \\
    \midrule
   \multirow{2}{*}{\textbf{CMU}}
       & JPE (cm)  & 0.75  & 0.61  & \textbf{0.41} \\
            & JOE ($^\circ$)   & 3.34  & 3.16  & \textbf{2.52} \\
     \cmidrule{1-5}
            \multirow{2}{*}{\parbox{1cm}{\centering \textbf{CMU- \\ Occlu}}} 
            & JPE (cm) & 0.87 & 1.09 & \textbf{0.46} \\
            & JOE ($^\circ$)  &3.55  & 5.13 & \textbf{2.60}\\
   \bottomrule
  \end{tabular}
  }
\vspace{2pt}
\caption{Ablation studies of our method.}
\label{tab:ablation}
\end{table}

\begin{table}[t]
  \centering
  \renewcommand{\arraystretch}{1.2}
  \resizebox{0.9\linewidth}{!}{
  \begin{tabular}{cccccc}
   \toprule
   \multirow{2}{*}{}& \multirow{2}{*}{}&  \multirow{2}{*}{\parbox{1cm}{\centering MoCap- \\ Solver}} & \multirow{2}{*}{\parbox{1cm}{\centering Local \\ MoCap}}& \multirow{2}{*}{RoMo} & \multirow{2}{*}{\parbox{1cm}{\centering Open \\ MoCap}} \\ 
   &&&&&
    \\
    \midrule
   \multirow{2}{*}{\textbf{CMU}} 
       & JPE (cm)    & 5.50  & 1.93 &  1.46   &\textbf{0.40} \\
    & JOE ($^\circ$)    & 10.03  & 4.86& 4.78    &\textbf{4.47} \\
     \cmidrule{1-6}
            \multirow{2}{*}{\parbox{1cm}{\centering \textbf{CMU- \\ Occlu}}} 
            & JPE (cm)  & 5.73 &1.41 &1.38  & \textbf{0.39}\\
            & JOE ($^\circ$)   & 10.21 &4.25 &4.22 &\textbf{4.10} \\
   \bottomrule
  \end{tabular}
  }
\vspace{2pt}
\caption{Comparison of methods trained on CMU and CMU-Occlu datasets, with evaluation conducted on SFU.}
\label{tab:sfu_test}
\end{table}

\begin{figure*}[t]
  \centering
   \includegraphics[width=1\linewidth]{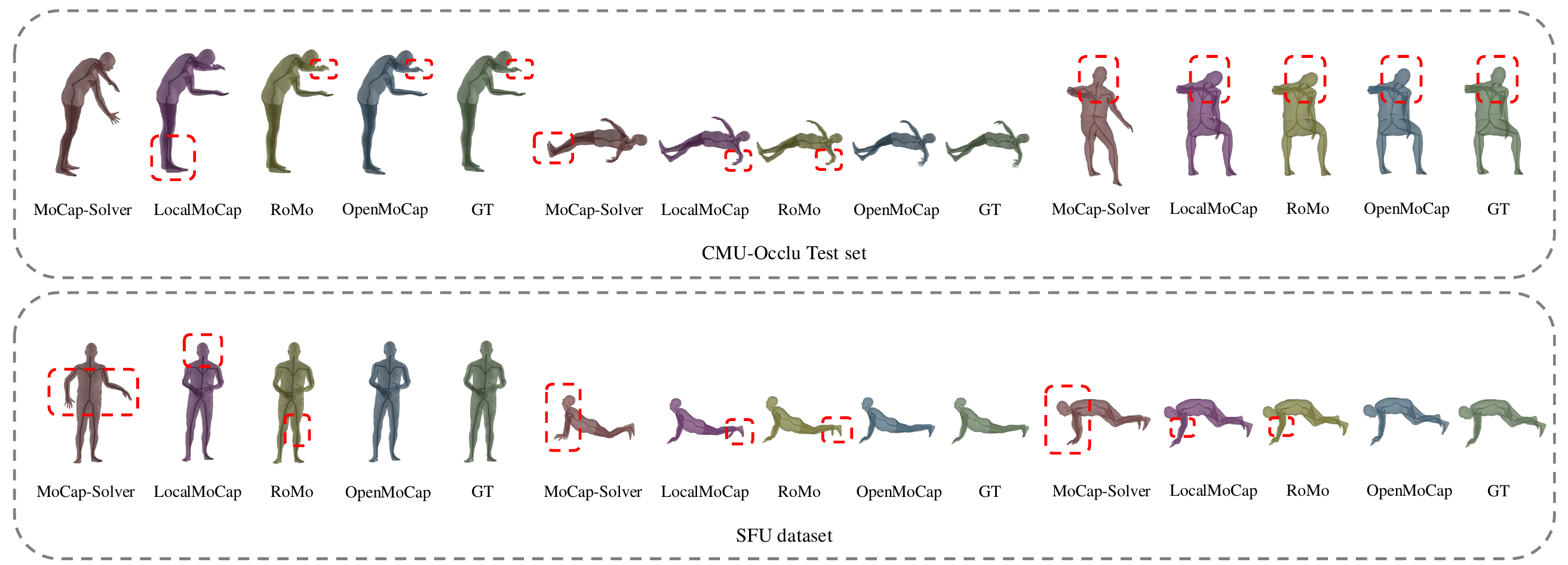}

   \caption{Qualitative evaluation of different models on CMU-Occlu test set and SFU dataset.}
   \label{fig:comparsion}
\end{figure*}

As occlusion increases, all methods show rising error. MoSh++ struggles with intra-sequence marker shifts, leading to high overall error. LocalMoCap and RoMo perform well under low occlusion but degrade sharply on the CMU-Occlu dataset due to their limited recovery strategies. In contrast, OpenMoCap models occluded markers as learnable parameters and employs a marker–joint chain inference mechanism to capture long-range dependencies, maintaining strong performance under realistic, high-occlusion conditions.

Fig.~\ref{fig:comparsion} shows visualization results. When key markers are missing, MoCap-Solver fails to align accurately, leading to large discrepancies from the ground truth. In contrast, OpenMoCap achieves more accurate reconstructions under severe occlusions. For example, in the bottom-right yoga pose with occluded abdominal and chest markers, it produces the closest reconstruction to the ground truth.

\subsection{Ablation Studies}
To evaluate component effectiveness, we conducted two ablation studies. In the first, we removed the marker–joint chain and directly estimated joint positions from visible markers, predicting occluded ones afterward. As discussed in Sec.\ref{sec:chain}, this weakens spatial reasoning and leads to inaccurate marker recovery, which further degrades rotation estimation. As shown in Tab.\ref{tab:ablation}, both joint position and rotation errors increase under this setting.

In the second experiment, we merged the position solver and rotation solver into a single network while proportionally increasing its depth. In the CMU-Occlu test set, certain sequences contain fewer than three visible key markers after occlusion. While the method performs satisfactorily on the CMU dataset, it requires the alignment process to be performed during data preprocessing, which can result in a notable degradation in performance on the CMU-Occlu dataset.

\begin{figure}[t]
  \centering
   \includegraphics[width=0.9\linewidth]{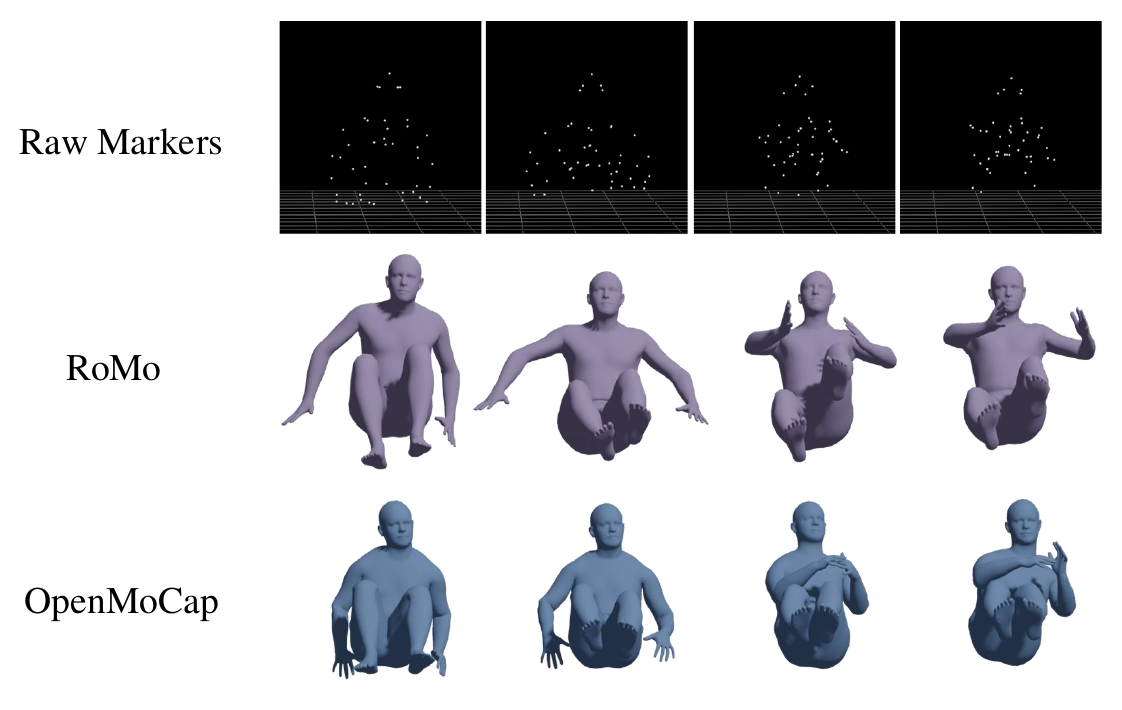}

   \caption{Qualitative evaluation of different models on real MoCap of a Russian twist.}
   \label{fig:exp_application}
\end{figure}

\begin{table}[t]
  \centering
  \renewcommand{\arraystretch}{1.2}
  \resizebox{1\linewidth}{!}{
  \begin{tabular}{ccccc}
   \toprule
 & Mocap-Solver & LocalMoCap & RoMo & OpenMoCap\\
    \midrule
       JPE (cm)    & 7.68  & 4.92 &  4.73   & \textbf{1.01} \\
    JOE ($^\circ$)   & 28.83 &16.73 &15.98 &\textbf{12.68} \\
   \bottomrule
  \end{tabular}
  }
  \vspace{2pt}
  \caption{Comparison of different models on the processed MOYO dataset.}
  
\label{tab:moyo}
\end{table}

\subsection{Dataset Analysis}
To evaluate the effectiveness of the dataset in supporting real-world deployment of pre-trained models, we train each method separately on the CMU and CMU-Occlu datasets and test them on the real-world MoCap dataset SFU. 

In the CMU dataset, complex motions often involve realistic occlusions around the waist and abdomen. As shown in Tab.~\ref{tab:sfu_test}, models like MoCap-Solver, which rely on marker visibility, struggle under such conditions, as misaligned inputs degrade training. In contrast, models with inherent occlusion robustness benefit from training on CMU-Occlu and perform better in real-world scenarios.

\subsection{Application}


We conducted the real-world experiment using MoSen MoCap system and tested the performance of different models. The actor performed a Russian twist, which involves four stages: sitting down, raising the legs, lifting the arms, and twisting the torso. As shown in Fig.\ref{fig:exp_application}, the raw markers represent the captured visible markers. Significant occlusions occurred in the abdomen, leg, and hip regions. We compared the inference results of the SOTA method and OpenMoCap. In contrast to RoMo, OpenMoCap robustly reconstructed the motion and successfully completed the solving process.

\subsection{Generalization Study}
To further explore the generalization ability of OpenMoCap, we processed the MOYO dataset~\cite{tripathi2023ipman}, which contains diverse and challenging yoga poses and strong self-occlusion, fine-tuned the models on this new domain and report the performance in Tab.\ref{tab:moyo}. The results indicate that OpenMoCap achieves a significant performance advantage compared to existing approaches.

\vspace{-2pt}
\section{Conclusion}
\label{sec:conclusion}
In this paper, we conduct an in-depth analysis of two key factors that lead to performance degradation when deploying existing models in real-world MoCap environments, and propose targeted solutions for each. First, we introduce the CMU-Occlu dataset, which incorporates more realistic marker occlusion patterns, thereby improving the distributional alignment between synthetic training data and real-world test scenarios. Second, we propose the OpenMoCap solver, which establishes strong long-range dependencies between markers through a marker-joint inference mechanism. Experimental results demonstrate that CMU-Occlu enhances model generalization, while OpenMoCap achieves robust motion solving under diverse occlusion conditions, surpassing SOTA performance.



\begin{acks}
We sincerely thank the MobiSense group and anonymous reviewers for their insightful comments. This work is supported in part by the National Key Research Plan under grant No. 2021YFB2900100, the NSFC under grant No. 62372265, No. 62302254, and No. 62402276.
\end{acks}

\bibliographystyle{ACM-Reference-Format}
\balance
\bibliography{main}


\end{document}